\documentclass[conference]{IEEEtran}
\IEEEoverridecommandlockouts
\usepackage{cite}
\usepackage{amsmath,amssymb,amsfonts}
\usepackage{graphicx}
\usepackage{textcomp}
\usepackage{xcolor}

\usepackage{algorithm}
\usepackage[algo2e]{algorithm2e}

\def\BibTeX{{\rm B\kern-.05em{\sc i\kern-.025em b}\kern-.08em
    T\kern-.1667em\lower.7ex\hbox{E}\kern-.125emX}}
\begin{document}

\title{
Novel Saliency Analysis for the Forward-Forward Algorithm  \\
}

\author{\IEEEauthorblockN{1\textsuperscript{st} Mitra Bakhshi}
\IEEEauthorblockA{\textit{Department of Civil Engineering} \\
\textit{Sapienza University of Rome}\\
 Rome, Italy \\
bakhshi.2058909@studenti.uniroma1.it}
}

\maketitle

\begin{abstract}
Incorporating the Forward-Forward algorithm into neural network training represents a transformative shift from traditional methods, introducing a dual-forward mechanism that streamlines the learning process by bypassing the complexities of derivative propagation. This method is noted for its simplicity and efficiency and involves executing two forward passes—the first with actual data to promote positive reinforcement, and the second with synthetically generated negative data to enable discriminative learning. Our experiments confirm that the Forward-Forward algorithm is not merely an experimental novelty but a viable training strategy that competes robustly with conventional multi-layer perceptron (MLP) architectures. To overcome the limitations inherent in traditional saliency techniques, which predominantly rely on gradient-based methods, we developed a bespoke saliency algorithm specifically tailored for the Forward-Forward framework. This innovative algorithm enhances the intuitive understanding of feature importance and network decision-making, providing clear visualizations of the data features most influential in model predictions. By leveraging this specialized saliency method, we gain deeper insights into the internal workings of the model, significantly enhancing our interpretative capabilities beyond those offered by standard approaches. Our evaluations, utilizing the MNIST and Fashion MNIST datasets, demonstrate that our method performs comparably to traditional MLP-based models.
\end{abstract}

\section{Introduction}

Artificial Intelligence (AI) and Machine Learning (ML) have revolutionized various sectors, including industry, public services, and society at large. The advent of deep learning (DL) has been particularly transformative, enabling AI systems to perform tasks such as image and speech recognition at a level that often matches or surpasses human capabilities \cite{mnih2015human, young2018recent}. Central to the success of DL is the backpropagation (BP) algorithm, which drives learning by iteratively adjusting the weights of neural networks to minimize prediction errors. Despite its effectiveness, BP's computational intensity and its lack of alignment with biological learning mechanisms have spurred the search for alternative methods that can function in resource-constrained environments while more closely mirroring natural learning processes \cite{lecun2015deep, sejnowski2020unreasonable}. In response to these challenges, research has increasingly focused on developing approaches that approximate the efficiency of BP while mitigating its limitations. Self-supervised learning, which enables models to learn from data without explicit labels, represents one such advancement. Additionally, algorithms like Feedback Alignment and Direct Feedback Alignment address BP's spatial non-locality and weight transport issues, offering promising alternatives \cite{lillicrap2016random, nokland2016direct}. Furthermore, techniques such as predictive coding and equilibrium propagation propose more biologically plausible models of learning by reducing the reliance on non-local computations \cite{marblestone2016toward, friston2010free}.

However, these alternative approaches often entail a trade-off between accuracy and biological plausibility, particularly when applied to complex datasets like ImageNet. The quest for algorithms that can achieve high performance without the limitations of end-to-end BP has led to the development of sophisticated models that incorporate local plasticity rules and auxiliary networks, albeit with increased complexity \cite{chen2020epidemiological}. The ongoing debate over the viability of BP as a model for brain learning has highlighted significant practical and biophysical challenges, further questioning the extent to which BP aligns with the brain's learning mechanisms. This skepticism has catalyzed interest in brain-inspired computing, which seeks to develop computational models that more accurately reflect the brain's learning processes \cite{hinton2022forward}. Among the promising avenues in this field are predictive coding and the forward-forward algorithm, which offer potential pathways to bridging the gap between artificial learning systems and the intricacies of biological neural networks. These approaches not only challenge our understanding of artificial systems but also deepen our insights into the cognitive processes that govern learning and intelligence in the natural world \cite{friston2010free}. As AI and ML continue to evolve, spearheaded by advancements in deep learning and convolutional neural networks, we find ourselves on the cusp of a new era of technological innovation. Yet, the pursuit of algorithms that harmonize efficiency, accuracy, and biological fidelity remains ongoing, driving the exploration of alternatives to traditional models like backpropagation. In this context, saliency maps have emerged as a crucial tool for enhancing the interpretability of deep neural networks (DNNs). These maps provide a visual representation of the elements that significantly influence a model's decisions, thereby offering insights into the internal workings of DNNs. By assigning scores to different input features based on their impact on the model's output, gradient-based saliency techniques help to clarify the decision-making processes within the network \cite{selvaraju2017grad, shrikumar2017learning}. This enhanced transparency is critical in applications where interpretability is essential, such as healthcare, neuroscience, financial services, and autonomous vehicle technology \cite{caruana2015intelligible, li2018tell}. Beyond improving trust in model predictions, saliency maps play a vital role in model refinement and troubleshooting, driving both the theoretical and practical advancements in deep learning.

In our investigation, we adopt the forward-forward (FF) algorithm, specifically tailored for multilayer perceptrons (MLP). We introduce a novel saliency mapping technique adapted for the FF algorithm, marking a departure from traditional methods that rely on backpropagation. By eliminating the use of backpropagation, our algorithm offers a fresh perspective on generating saliency maps within the FF framework, thereby enhancing the interpretability of neural networks without the computational complexities and limitations associated with backpropagation \cite{hinton2022forward}. This development represents a significant advance toward more efficient and transparent deep learning methodologies.

\section{Related Work}

Recent advances in deep learning underscore the efficacy of stochastic gradient descent applied to models with expansive parameter spaces and significant data volumes. Central to this process is backpropagation \cite{rumelhart1986learning}, which computes gradients essential for training. Despite its widespread adoption, the biological plausibility of backpropagation remains a topic of debate. Scholars question whether similar mechanisms exist in the human brain for synaptic weight adjustment \cite{lillicrap2020backpropagation, guerguiev2017towards}. Yet, there is no conclusive evidence that the cortex employs mechanisms akin to error derivative propagation or reverses computation phases. Further complicating this hypothesis, the brain's cortical connections form complex loops rather than the hierarchical structures typical of backpropagation, posing challenges for its biological feasibility, especially in processing sequential data \cite{lillicrap2020backpropagation}. Additionally, the brain’s ongoing processing of sensory information without pausing for error correction suggests an alternative, dynamic learning strategy that adjusts synapses in real-time, contrasting sharply with the sequential nature of backpropagation. The opacity introduced in the forward pass of deep neural networks (DNNs) also complicates gradient derivation, necessitating alternative models that can manage non-differentiable elements \cite{guerguiev2017towards}. Deep learning has profoundly impacted various sectors by enabling the extraction of complex patterns from extensive datasets, leading to precise predictions and informed decision-making \cite{lecun2015deep}. This issue is critical in high-stakes domains such as healthcare, finance, and autonomous driving, where decisions must be both accurate and interpretable \cite{caruana2015intelligible, li2018tell, karkehabadihlgm}. Addressing these challenges, significant research efforts have focused on developing methods to enhance DNN interpretability. These include using saliency maps to identify influential input features, although their effectiveness can be diminished by noise and other artifacts \cite{selvaraju2017grad, shrikumar2017learning}. Techniques such as SmoothGrad and Integrated Gradients have been introduced to refine these visualizations by averaging the effects of noise or modifying gradient functions to offer deeper insights into decision processes \cite{smilkov2017smoothgrad, ancona2017towards, hassanpour2024overcoming}. SMOOT \cite{karkehabadi2024smoot} builds upon these saliency-based methods by introducing a novel approach that optimizes the number of masked images during training, significantly improving both model accuracy and the prominence of salient features. This work demonstrates how careful adjustment of the masking strategy can prevent information loss and enhance interpretability without sacrificing predictive performance. Research has shown that there is often a trade-off between noise and power consumption in various systems \cite{ghiasi2022simple}, highlighting the importance of optimizing both to improve system performance. Similarly, in the context of the Internet of Things (IoT), especially in underwater environments, efficient data routing is vital for minimizing resource consumption while ensuring reliable communication. Recent work \cite{karkehabadi2024optimizing} has introduced a method that optimizes network performance by integrating multi-criteria decision-making with uncertainty weights, thereby enhancing underwater IoT communications. The pursuit of robust interpretability not only aims to clarify model decisions but also seeks to understand the representations learned by DNNs. Exploring methods like network distillation into interpretable models, such as soft decision trees, presents promising directions for enhancing transparency in machine learning applications \cite{frosst2017distilling}.

The Forward-Forward Algorithm represents a novel approach in the domain of neural network learning, particularly when dealing with unidentified nonlinearities. It eliminates the need for traditional reinforcement learning by allowing networks to learn directly from sequential data without storing neural activities or interrupting error propagation. This algorithm operates comparably to backpropagation but does not require an in-depth understanding of the forward computational steps. However, it marginally lags behind backpropagation in terms of speed when tested across various toy problems, suggesting that its utility may be limited in scenarios where computational resources are ample~\cite{jabri1992weight}. Despite these limitations, the Forward-Forward Algorithm holds potential advantages for modeling cortical learning processes and could be better suited for use in low-power analog hardware, avoiding the complexities of reinforcement learning. Its foundational concept is influenced by mechanisms from Boltzmann machines~\cite{hinton1986learning} and Noise Contrastive Estimation~\cite{gutmann2010noise}, employing two forward passes instead of the traditional forward and backward passes. Each pass targets distinct datasets with diametrically opposed objectives: the first, or 'positive' pass, processes real data to increase a 'goodness' measure across each layer, while the second, or 'negative' pass, uses synthetic or 'negative' data to decrease this measure. The algorithm's efficacy is gauged through two specific measures of 'goodness'—the sum of squared neural activities and its inverse. This dual-process aims to refine the model's ability to classify input vectors as positive or negative data based on a logistic function, \(\sigma\), applied to the difference between the goodness measure and a predefined threshold, \(\theta\):

\begin{equation}
p(\text{positive}) = \sigma\left(\sum_{j} (y_{j}^2) - \theta\right)
\end{equation}

\begin{figure}[hbt!]
    \centering
    \includegraphics[width=0.9\columnwidth]{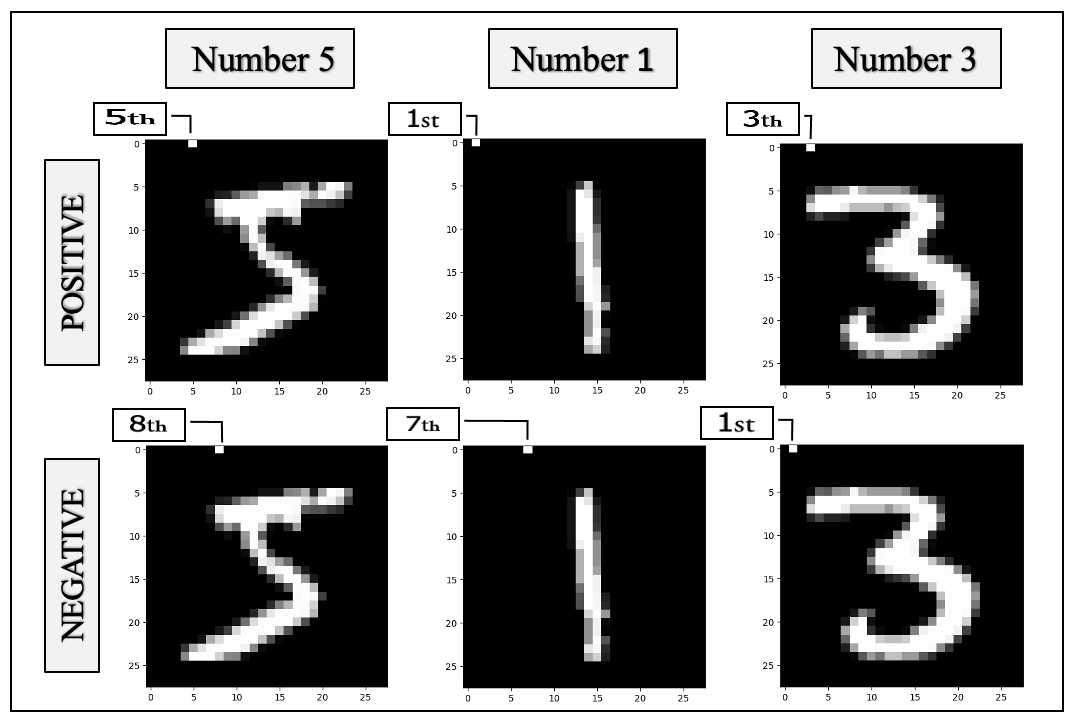}
    \vspace{-10pt}
    \caption{MNIST images contain a black border. If we replace
the first 10 pixels by one of N representations of the label. To develop positive and negative samples for the forward-forward algorithm, we change these N pixels. For a positive sample, set the tensor index corresponding to the class to 1 and all other indices to 0. This denotes the class's presence. Conversely, for a negative sample, choose a random index not matching the actual class, set this to 1, and all other indices to 0, indicating the absence of the class. This method allows the algorithm to distinguish between classes by clearly identifying which class is present (positive sample) and which is not (negative sample).}
    \label{Mnist}
\end{figure}

This strategic innovation highlights a shift toward more efficient and applicable neural property training methodologies, moving away from conventional learning paradigms and potentially offering a method better aligned with biological learning processes. Figure \ref{Mnist} illustrates the technique for generating negative and positive data in the forward-forward algorithm.

\begin{algorithm}[H]
\SetAlgoLined
\caption{Forward-Forward Algorithm }
\For{$\text{l} \in \text{model.layers}$}{
    \For{$\text{e} \in \text{MaxEpock}$}{
            \# \textcolor{blue}{Prepare pos and neg samples}\\
            $x_{e}, L_{e+} = get\_training\_sample(e)$\\            
            $L_{pos}, L_{neg} = \{C\{0\}\};$       \# concat C 0s   \\
            $L_{e-} = random(0, C, \! L_{e+})$;\\
            $L_{pos}[L_e]=1;$ \textcolor{blue}{change 0 in label position to 1} \\
            $L_{neg}[L_{ne}]=1;$ \textcolor{blue}{change 0 in a non-label position to 1}\\
            $x_{pos} = replace\_boarder(x_e, L_{pos})$ \\
            $x_{neg} = replace\_boarder(x_e, L_{neg})$ 

            \# \textcolor{blue}{Run pos and neg samples to target layer}\\
            $g_{\text{pos}} = \text{RunLayers}(0, l, x_{pos})$ \\
            $g_{\text{neg}} = \text{RunLayers}(0, l, x_{neg})$ \\
            \# \textcolor{blue}{Compute loss}\\
            $\text{loss} = \frac{1}{2}(\log(1 + e^{-g_{\text{pos}}}) + \log(1 + e^{g_{\text{neg}}}))$ \\
            \# \textcolor{blue}{Update weights}\\
            $W_{grad} = \text{one\_layer\_backpropagate}(loss);$ \\
            $\text{model.layer(l).weights\_update}(W_{grad});$
    }
}
\label{alg}
\end{algorithm}

The Forward-Forward Algorithm \ref{alg} is designed to optimize a neural network's performance through a systematic training process. It initializes by partitioning the available dataset into training and testing sets. The algorithm then iteratively adjusts the model's weights over a predefined number of epochs and iterations within each epoch. In each iteration, the algorithm identifies positive samples, where the class is correctly labeled, and negative samples, which are randomly chosen from non-matching classes. For both sets of samples, it computes the activations at each layer of the model.

The algorithm computes the gradients of the loss function, which is designed to penalize the model for incorrect classifications. Specifically, for positive samples (\(g_{\text{pos}}\)), the loss increases when the model's confidence in the correct classification is low. Conversely, for negative samples (\(g_{\text{neg}}\)), the loss increases when the model incorrectly classifies them as positive. The loss function is defined as \(\text{loss} = \left(\log(1 + e^{-g_{\text{pos}}}) + \log(1 + e^{g_{\text{neg}}})\right) / 2\), which combines the penalties for both types of errors in a manner that encourages the model to correctly classify both positive and negative samples with high confidence. Through backpropagation, the algorithm updates the model's weights based on the computed gradients, progressively reducing the classification error and enhancing the model's accuracy over time.

\section{Methodology}
\subsection{Accuracy Differential Saliency (ADS) Technique}

\begin{figure}[hbt!]
    \centering
    \includegraphics[width=1.01\columnwidth]{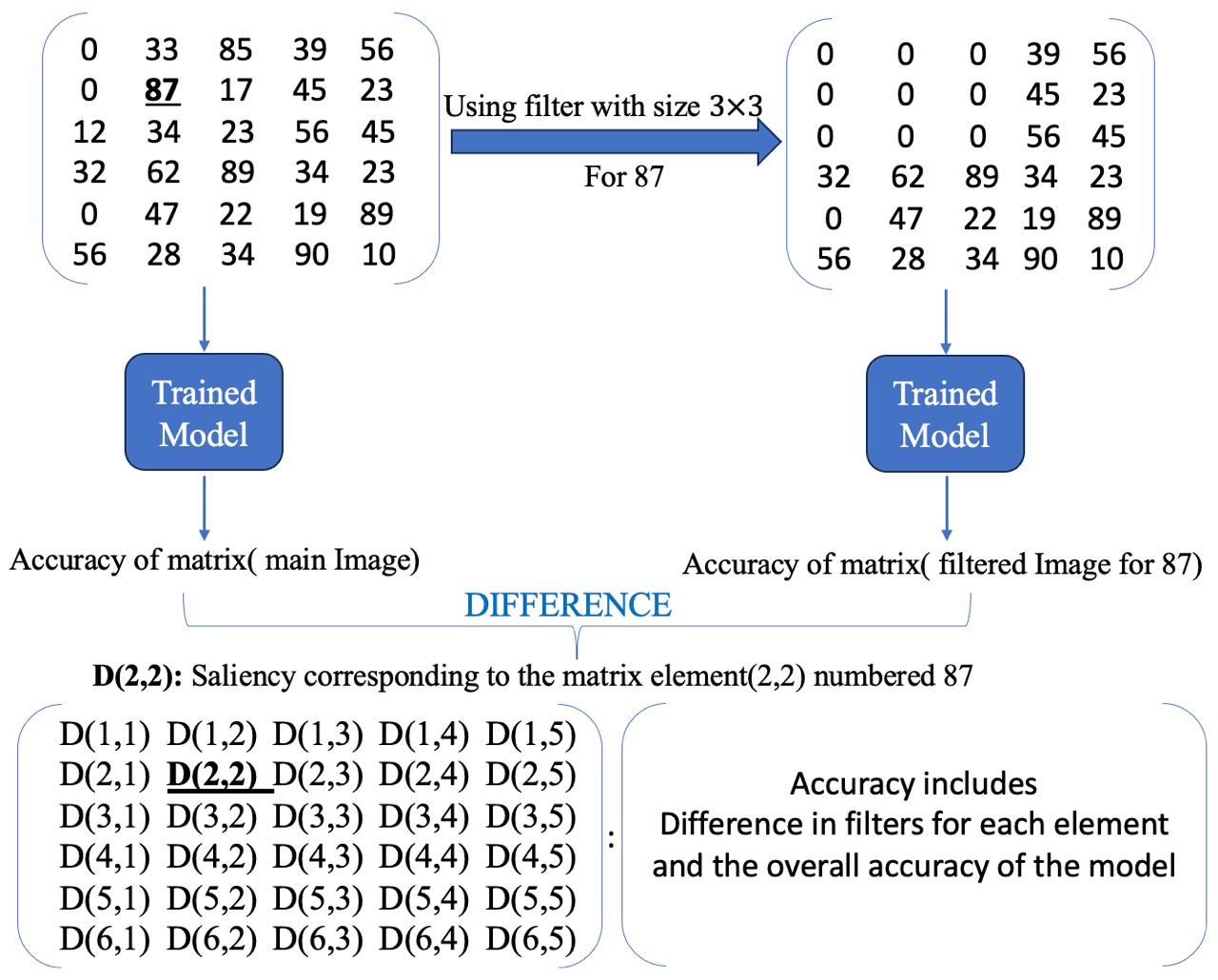}
    \vspace{1 pt}
    \caption{Plotting the loss for each layer in the forward-forward algorithm, where each layer learns independently, starting with the first layer, then proceeding to the second}
    \label{sali}
\end{figure}

In the context of saliency analysis, we employ an alternative approach due to the inherent limitations of the forward propagation algorithm, notably its lack of backpropagation as found in conventional neural networks. Our method focuses on quantifying the influence of individual pixels on the model's performance by systematically nullifying their contribution. This is achieved by applying a filter to the image, which iteratively moves across the entire image. For each position of the filter, pixels within its scope are set to zero, effectively removing their influence. Subsequently, we assess the model's accuracy without the contribution of these pixels. This process is repeated for every pixel position, with the pixel under the filter's center being the subject of analysis each time. By comparing the model's accuracy with and without the influence of each pixel, we derive a differential accuracy for each pixel. The aggregation of these differential accuracies forms a difference matrix, which serves as a visual representation of the impact each pixel has on the model's overall accuracy. This difference matrix thus provides insightful data on the saliency within the image, highlighting regions of particular importance to the model's decision-making process. figure \ref{sali}
shows the ADS Method.

\section{Experiments and Results}


In this section, we assess the performance of our method on the MNIST \cite{lecun1998mnist} and Fashion-MNIST \cite{xiao2017fashion} datasets. Both datasets consist of images with dimensions of \(28 \times 28\) pixels. These images were converted into a flattened format to form an input layer with 784 units.
All computational experiments were performed using an NVIDIA T4 GPU, equipped with 2,560 CUDA cores, 320 Tensor cores, and 16 GB of GDDR6 memory.

\section*{Architecture}
\begin{figure}[hbt!]
    \centering
    \includegraphics[width=1.01\columnwidth]{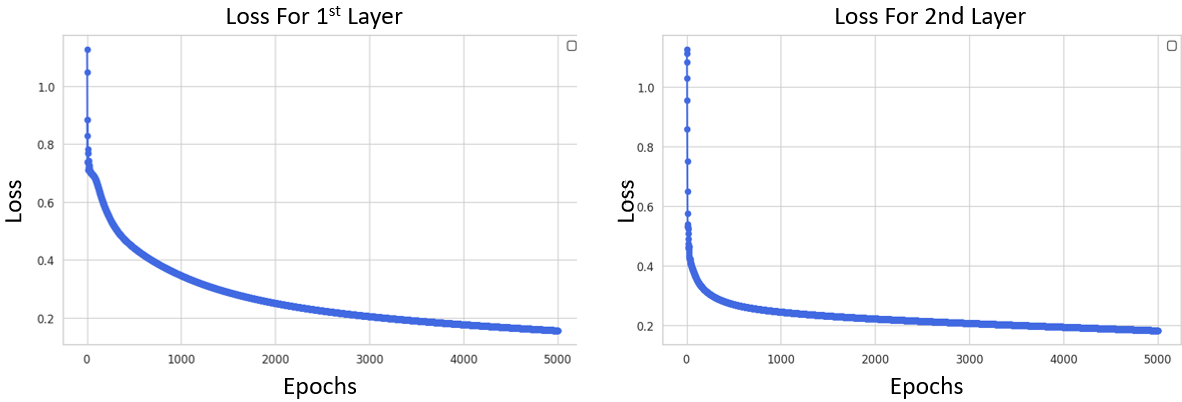}
    \caption{Plotting the loss for each layer in the forward-forward algorithm, where each layer learns independently, starting with the first layer, then proceeding to the second}
    \label{loss}
\end{figure}

We conducted several experiments to showcase the forward-forward algorithm in image classification. We used two linear layers for the MLP architecture with a hidden layer size of 500. The MLP model was trained for 5,000 epochs.
Figure \ref{loss} illustrates the loss associated with each layer. In the forward-forward algorithm, each layer learns independently, starting with the first layer and proceeding sequentially through the others.

\begin{figure}[hbt!]
    \centering
    \includegraphics[width=1\columnwidth]{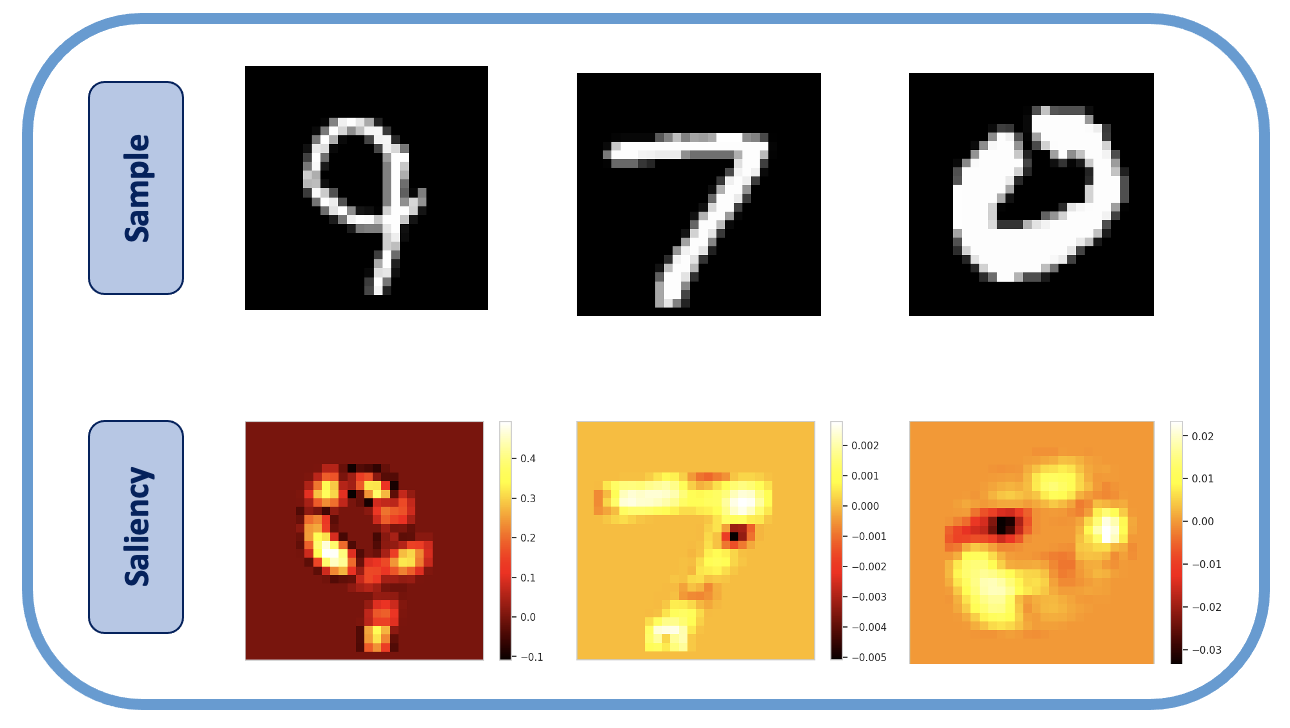}
    \vspace{-10 pt}
    \caption{ The Accuracy Differential Saliency (ADS) Technique, applied to the MNIST dataset, evaluates the impact of individual pixels on model accuracy by zeroing them out with a moving filter. This process generates a difference matrix by comparing the model's performance with and without each pixel's contribution, highlighting key areas affecting decision-making. The results are visualized by plotting this matrix over the image.}
    \label{mnistr}
\end{figure}

\begin{figure}[hbt!]
    \centering
    \includegraphics[width=1\columnwidth]{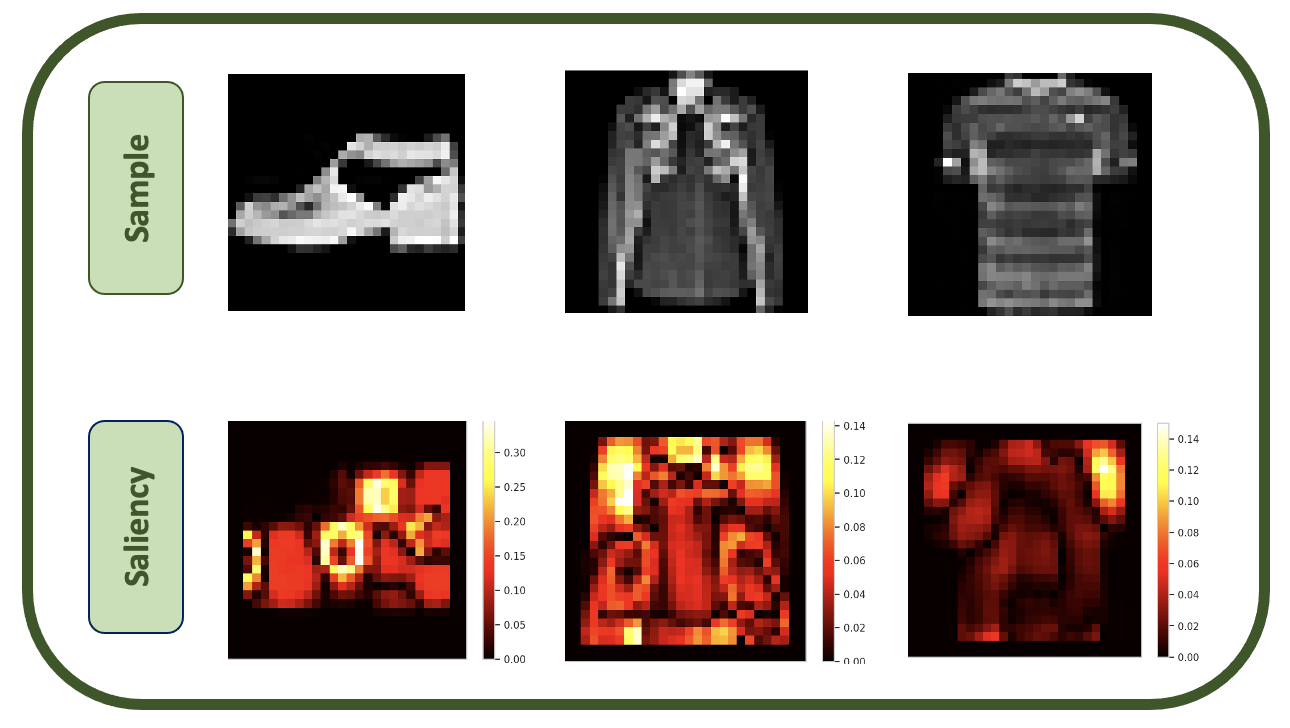}
    \vspace{-10 pt}
    \caption{ The Accuracy Differential Saliency (ADS) Technique, applied to the Fashion MNIST dataset, evaluates the impact of individual pixels on model accuracy by zeroing them out with a moving filter.}
    \label{fashionr}
\end{figure}

For our new technique, titled "The Accuracy Differential Saliency (ADS) Technique," we applied it to the MNIST dataset to assess the impact of individual pixels on model accuracy. This was achieved by systematically nullifying them using a moving filter. The technique creates a differential matrix that contrasts the model's performance with and without the contribution of each pixel, thereby identifying critical regions that influence decision-making. The outcomes are illustrated by overlaying this matrix onto the image. Figures \ref{mnistr} and \ref{fashionr} display these results for the MNIST and Fashion MNIST datasets.

\section{Conclusion}
Our study demonstrates the efficacy of the Forward-Forward algorithm in neural network training, marking a significant departure from traditional backpropagation methods. By integrating a specialized saliency algorithm tailored for this non-backpropagation approach, we enhance the interpretability of neural networks, offering a more intuitive understanding of feature importance and decision-making processes. Our evaluations using the MNIST and Fashion MNIST datasets show that this method not only performs on par with traditional multi-layer perceptron (MLP) architectures but also simplifies the training process. This approach opens new avenues for the development of efficient and interpretable training methods, setting a promising direction for future research in deep learning advancements that cater to both spatial and temporal data complexities.\\


\begin{thebibliography}{00}

\bibitem{mnih2015human}Mnih, V., Kavukcuoglu, K., Silver, D., Rusu, A., Veness, J., Bellemare, M., Graves, A., Riedmiller, M., Fidjeland, A., Ostrovski, G. \& Others Human-level control through deep reinforcement learning. {\em Nature}. \textbf{518}, 529-533 (2015)

\bibitem{young2018recent}Young, T., Hazarika, D., Poria, S. \& Cambria, E. Recent trends in deep learning based natural language processing. {\em Ieee Computational IntelligenCe Magazine}. \textbf{13}, 55-75 (2018)

\bibitem{lecun2015deep}LeCun, Y., Bengio, Y. \& Hinton, G. Deep learning. {\em Nature}. \textbf{521}, 436-444 (2015)




\bibitem{sejnowski2020unreasonable}Sejnowski, T. The unreasonable effectiveness of deep learning in artificial intelligence. {\em Proceedings Of The National Academy Of Sciences}. \textbf{117}, 30033-30038 (2020)

\bibitem{hinton2022forward}Hinton, G. The forward-forward algorithm: Some preliminary investigations. {\em ArXiv Preprint ArXiv:2212.13345}. (2022)



\bibitem{caruana2015intelligible}Caruana, R., Lou, Y., Gehrke, J., Koch, P., Sturm, M. \& Elhadad, N. Intelligible models for healthcare: Predicting pneumonia risk and hospital 30-day readmission. {\em Proceedings Of The 21th ACM SIGKDD International Conference On Knowledge Discovery And Data Mining}. pp. 1721-1730 (2015)

\bibitem{li2018tell}Li, K., Wu, Z., Peng, K., Ernst, J. \& Fu, Y. Tell me where to look: Guided attention inference network. {\em Proceedings Of The IEEE Conference On Computer Vision And Pattern Recognition}. pp. 9215-9223 (2018)


\bibitem{chen2020epidemiological}Chen, N., Zhou, M., Dong, X., Qu, J., Gong, F., Han, Y., Qiu, Y., Wang, J., Liu, Y., Wei, Y. \& Others Epidemiological and clinical characteristics of 99 cases of 2019 novel coronavirus pneumonia in Wuhan, China: a descriptive study. {\em The Lancet}. \textbf{395}, 507-513 (2020)
\bibitem{lillicrap2016random}Lillicrap, T., Cownden, D., Tweed, D. \& Akerman, C. Random synaptic feedback weights support error backpropagation for deep learning. {\em Nature Communications}. \textbf{7}, 13276 (2016)
\bibitem{nokland2016direct}Nøkland, A. Direct feedback alignment provides learning in deep neural networks. {\em Advances In Neural Information Processing Systems}. \textbf{29} (2016)

\bibitem{marblestone2016toward}Marblestone, A., Wayne, G. \& Kording, K. Toward an integration of deep learning and neuroscience. {\em Frontiers In Computational Neuroscience}. \textbf{10} pp. 94 (2016)
\bibitem{friston2010free}Friston, K. The free-energy principle: a unified brain theory?. {\em Nature Reviews Neuroscience}. \textbf{11}, 127-138 (2010)


\bibitem{smilkov2017smoothgrad}Smilkov, D., Thorat, N., Kim, B., Viégas, F. \& Wattenberg, M. Smoothgrad: removing noise by adding noise. {\em ArXiv Preprint ArXiv:1706.03825}. (2017)
\bibitem{selvaraju2017grad}Selvaraju, R., Cogswell, M., Das, A., Vedantam, R., Parikh, D. \& Batra, D. Grad-cam: Visual explanations from deep networks via gradient-based localization. {\em Proceedings Of The IEEE International Conference On Computer Vision}. pp. 618-626 (2017)
\bibitem{shrikumar2017learning}Shrikumar, A., Greenside, P. \& Kundaje, A. Learning important features through propagating activation differences. {\em International Conference On Machine Learning}. pp. 3145-3153 (2017)
\bibitem{ancona2017towards}Ancona, M., Ceolini, E., Öztireli, C. \& Gross, M. Towards better understanding of gradient-based attribution methods for deep neural networks. {\em ArXiv Preprint ArXiv:1711.06104}. (2017)


\bibitem{ghiasi2022simple}Ghiasi, M., Moezzi, M. \& Kashi, A. A Simple Low Phase Noise Class-F LC Oscillator. {\em Circuits, Systems, And Signal Processing}. \textbf{41}, 3041-3049 (2022)
\bibitem{frosst2017distilling}Frosst, N. \& Hinton, G. Distilling a neural network into a soft decision tree. {\em ArXiv Preprint ArXiv:1711.09784}. (2017)

\bibitem{karkehabadi2024optimizing}Karkehabadi, A., Bakhshi, M. \& Razavian, S. Optimizing Underwater IoT Routing with Multi-Criteria Decision Making and Uncertainty Weights. {\em ArXiv Preprint ArXiv:2405.11513}. (2024)





\bibitem{jabri1992weight}Jabri, M. \& Flower, B. Weight perturbation: An optimal architecture and learning technique for analog VLSI feedforward and recurrent multilayer networks. {\em IEEE Transactions On Neural Networks}. \textbf{3}, 154-157 (1992)

\bibitem{karkehabadihlgm}Karkehabadi, A., Latibari, B., Homayoun, H. \& Sasan, A. HLGM: A Novel Methodology For Improving Model Accuracy Using Saliency-Guided High and Low Gradient Masking. {\em The 14th International Conference On Information Science And Technology}.

\bibitem{hassanpour2024overcoming}Hassanpour, J., Srivastav, V., Mutter, D. \& Padoy, N. Overcoming Dimensional Collapse in Self-supervised Contrastive Learning for Medical Image Segmentation. {\em ArXiv Preprint ArXiv:2402.14611}. (2024)


\bibitem{hinton1986learning}Hinton, G. \& Others Learning distributed representations of concepts. {\em Proceedings Of The Eighth Annual Conference Of The Cognitive Science Society}. \textbf{1} pp. 12 (1986)
\bibitem{gutmann2010noise}Gutmann, M. \& Hyvärinen, A. Noise-contrastive estimation: A new estimation principle for unnormalized statistical models. {\em Proceedings Of The Thirteenth International Conference On Artificial Intelligence And Statistics}. pp. 297-304 (2010)


\bibitem{karkehabadi2024smoot}Karkehabadi, A., Homayoun, H. \& Sasan, A. SMOOT: Saliency guided mask optimized online training. {\em 2024 IEEE 17th Dallas Circuits And Systems Conference (DCAS)}. pp. 1-6 (2024)




\bibitem{xiao2017fashion}Xiao, H., Rasul, K. \& Vollgraf, R. Fashion-mnist: a novel image dataset for benchmarking machine learning algorithms. {\em ArXiv Preprint ArXiv:1708.07747}. (2017)



\bibitem{lillicrap2020backpropagation}Lillicrap, T., Santoro, A., Marris, L., Akerman, C. \& Hinton, G. Backpropagation and the brain. {\em Nature Reviews Neuroscience}. \textbf{21}, 335-346 (2020)
\bibitem{guerguiev2017towards}Guerguiev, J., Lillicrap, T. \& Richards, B. Towards deep learning with segregated dendrites. {\em Elife}. \textbf{6} pp. e22901 (2017)


\bibitem{lecun1998mnist}LeCun, Y. The MNIST database of handwritten digits. {\em Http://yann. Lecun. Com/exdb/mnist/}. (1998)


\bibitem{rumelhart1986learning}Rumelhart, D., Hinton, G. \& Williams, R. Learning representations by back-propagating errors. {\em Nature}. \textbf{323}, 533-536 (1986)




\end{thebibliography}
\end{document}